\documentclass[letterpaper, 10 pt, conference]{ieeeconf}  

\IEEEoverridecommandlockouts                              
\usepackage[pdftex]{graphicx}
\usepackage{amsmath}
\usepackage{subfigure}
\usepackage{tabularx}
\usepackage{cite}
\usepackage{color}
\usepackage{stfloats}
\usepackage[bookmarks=true,colorlinks,linkcolor=black,citecolor=black,urlcolor=black]{hyperref}
\newtheorem{theorem}{Theorem}

\graphicspath{{Graphics/}}

\title{\LARGE \bf Variable Stiffness Control with Strict Frequency Domain Constraints for Physical Human-Robot Interaction}

\author{Wulin Zou, Pu Duan, Yawen Chen, Ningbo Yu and Ling Shi \\
	\thanks{W. Zou, Y. Chen and L. Shi are with the Department of Electronic and Computer Engineering, Hong Kong University of Science and Technology, Hong Kong, China. Email:wzouab@connect.ust.hk, ychenga@connect.ust.hk, eesling@ust.hk.}
	\thanks{P. Duan is with Xeno Dynamics Co., Ltd, Shenzhen, China. Email: duanpu@xeno.com.}
	\thanks{N. Yu is with the Institute of Robotics and Automatic Information Systems, and Tianjin Key Laboratory of Intelligent Robotics, Nankai University, Tianjin, China. Email: nyu@nankai.edu.cn. Corresponding author.}%
	\thanks{This work has been supported by the National Natural Science Foundation of China (61873135, U1713223, U1913208) and the Hong Kong ITF Fund GHP/001/18SZ.}%
}

\begin{document}

\maketitle
\thispagestyle{empty}
\pagestyle{empty}

\begin{abstract}

Variable impedance control is advantageous for physical human-robot interaction to improve safety, adaptability and many other aspects. This paper presents a gain-scheduled variable stiffness control approach under strict frequency-domain constraints. Firstly, to reduce conservativeness, we characterize and constrain the impedance rendering, actuator saturation, disturbance/noise rejection and passivity requirements into their specific frequency bands. This relaxation makes sense because of the restricted frequency properties of the interactive robots. Secondly, a gain-scheduled method is taken to regulate the controller gains with respect to the desired stiffness. Thirdly, the scheduling function is parameterized via a nonsmooth optimization method. Finally, the proposed approach is validated by simulations, experiments and comparisons with a gain-fixed passivity-based PID method.

\end{abstract}

\section{Introduction}

As robots have extensively evolved and have been designed to intimately coexist and cooperate with humans to execute various tasks in unstructured environment, safety and adaptability come to be essential considerations. During physical interaction between the robot and human or environment, the robot should present intrinsic compliant behavior and the ability to vary its output impedance for better adaptation to different tasks or dynamical environment.

Since impedance or compliance can be defined by the dynamic relationship between the output force and motion of the robot's end-effector, hybrid force/position control, impedance or admittance control strategies have been developed and widely applied to regulate the desired behavior. However, the hybrid force/position control is better suitable for well-structured environment, while impedance or admittance control can deal with both structured and unstructured environment. Impedance control, proposed by Hogan~\cite{Hogan1984}, has been intensively investigated in numerous applications, such as manipulators \cite{Roveda2015ICINCO}, teleoperation robots \cite{Ferraguti2015TRO}, haptic devices \cite{Ryu2008CTA}, rehabilitation and assistive robots \cite{LiXiang2018Automatica,Zhang2019RAL}, etc.

Variable impedance control also attracted growing research effort. In \cite{Ryu2008CTA,Dimeas2016TH,Landi2017IROS}, several adaptation strategies of admittance parameters were proposed to avoid unstable behavior by detecting high-frequency oscillations during interaction. Nevertheless, these methods were mainly focused on interaction stability, and used the variation of the impedance/admittance parameters to suppress unstable behavior. In \cite{Lee2008IFAC,Ficuciello2015TRO}, the target impedance was adjusted to achieve more accurate interaction control. In \cite{Tsumugiwa2002ICRA,Duchaine2007WHC,Lecours2012ICRA}, variable impedance controllers were applied in diverse human-robot cooperative tasks to achieve either stiff or compliant actuation. In \cite{Buchli2011IJRR,Dimeas2015IROS,Zhang2017TIE}, learning or neural network based methods were employed to realize variable impedance rendering. In \cite{Ferraguti2013ICRA,Kastritsi2018RAL}, energy tank based approach was used to guarantee passivity for variable impedance control. Variable impedance actuators are also able to change output impedance using additional mechanical units~\cite{Vanderborght2013RAS}, but are not in the scope of this paper and will not be discussed here. 

For variable impedance control in real applications, rendering accuracy, actuator saturation, energy consumption, robustness against disturbance and noise, and stability/passivity should all be considered into the design process. However, this is challenging and there is limited work regarding variable impedance/admittance control that can meet all these constraints simultaneously. In addition, little work on impedance/admittance control took into consideration the fact that human-robot interaction is intrinsically bounded into the low frequency range. Thus, the resulted controllers have to meet the requirements in full frequency domain, and are inevitably conservative. Haninger and Tomizuka \cite{Haninger2018Tmech} discussed and defined the relaxed passivity by frequency-domain inequality. However, it mainly focused on the influence of the PID-based force-loop controller gains on the relaxed passivity, and did not intend to design a passivity controller, or solve the problem of variable impedance control with multiple frequency-domain constraints.

In our previous work \cite{Yu2018MSSP,Yu2019CTA}, we improved stiffness rendering performance by restricting constraints on accuracy, actuator limitation, disturbance attenuation and noise rejection into their specific frequency ranges. Strict full frequency-domain passivity was guaranteed for constant stiffness control in \cite{Yu2018MSSP}. In this paper, we extend the methods to variable stiffness control by scheduling the gains, and the desired stiffness can be smoothly varied online while all the performance constraints are guaranteed. The passivity constraint is relaxed in such a way that it is only required for the low frequency range. These are the two key differences between this paper and the previous two \cite{Yu2018MSSP,Yu2019CTA}. Simulations and experiments are performed to validate the methods and compared with a gain-fixed passivity-based PID method. Thus, the main contributions of this paper are summarized as follows.
\begin{enumerate}
	\item The conventional full frequency-domain passivity is relaxed and only restricted in the low frequency based on the fact that the robots usually have restricted frequency properties. Along with other four frequency-domain constraints regarding accuracy, actuator saturation, disturbance attenuation and noise rejection, it can be solved by a nonsmooth optimization method.
	\item A gain-scheduled variable stiffness control approach with frequency-domain constraints is proposed for physical human-robot interaction to achieve stable and smooth stiffness transition.
\end{enumerate}

\section{The Variable Impedance Control Problem}
\label{Section2}

\begin{figure}[!htbp]
	\centering
	\includegraphics[width=0.5\columnwidth]{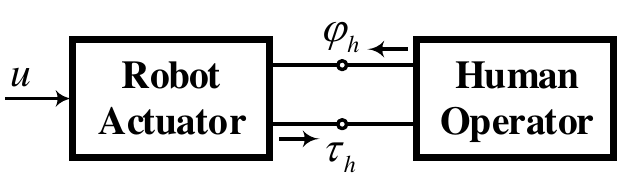}
	\caption{Physical interaction between the robot actuator and human operator.}
	\label{fig_HRI}
\end{figure}

The diagram of physical human-robot interaction is illustrated in Fig.~\ref{fig_HRI}, where $\varphi_h$ and $\tau_h$ are the interactive motion and torque between the human operator and robot, $u$ denotes the control signal of the actuator. The following transfer relationships can be obtained,
\begin{equation}
	G_1(s)=\frac{\tau_h(s)}{u(s)},~G_2(s)=\frac{\tau_h(s)}{\varphi_h(s)}.
\end{equation}

For the interactive system, the impedance $Z(s)$ can be defined as the transfer function from the interactive motion to the interactive torque, i.e., $Z(s)=\displaystyle \frac{\tau_h(s)}{-\varphi_h(s)}$,
where, the minus denotes the opposite direction of motion and torque. For the open-loop system, the impedance is $-G_2(s)$.

\begin{figure}[!htbp]
	\centering
	\includegraphics[width=0.55\columnwidth]{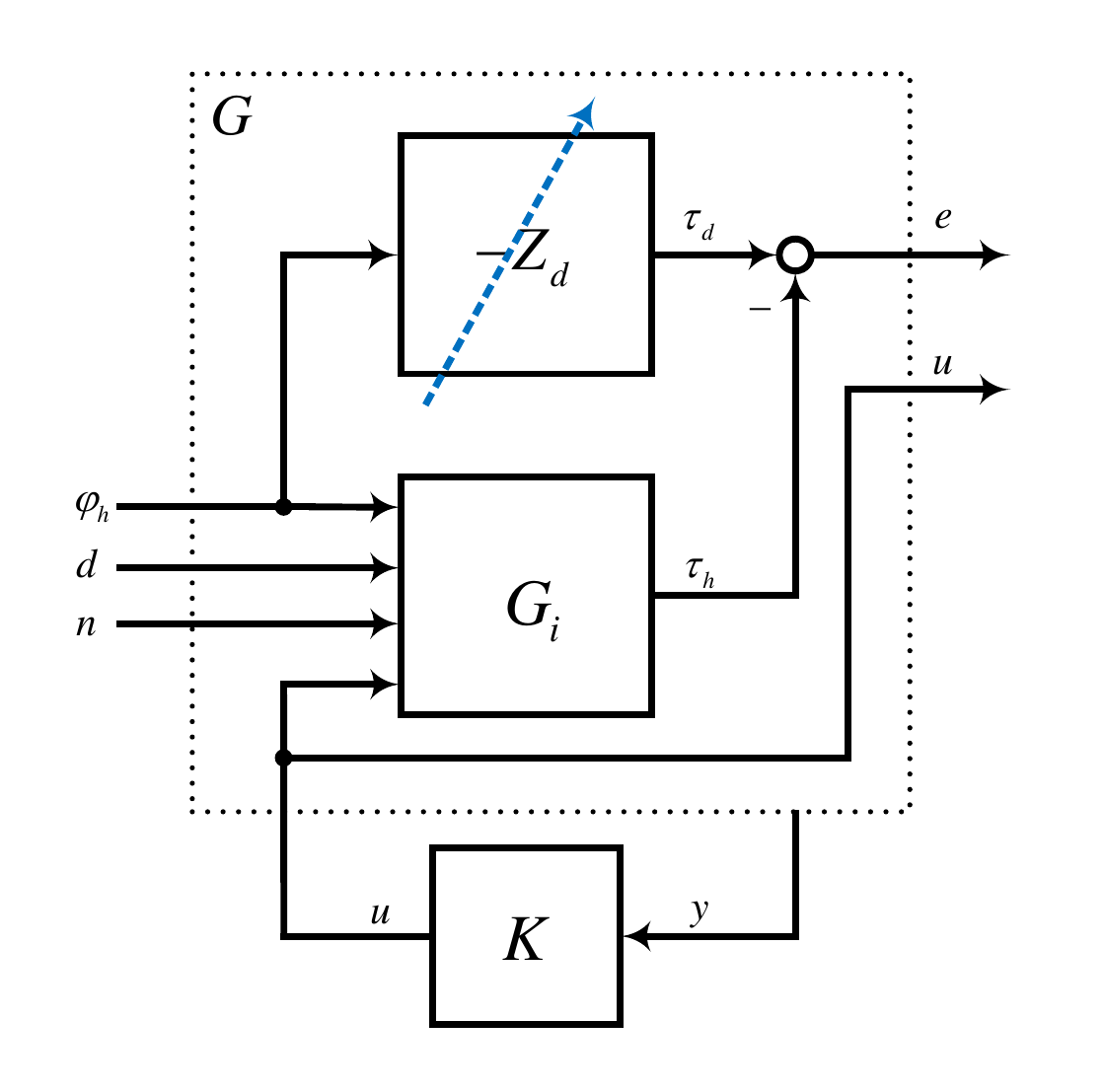}
	\caption{The impedance matching framework.}
	\label{fig_impedance_matching}
\end{figure}

The objective of impedance control is to shape the closed-loop impedance to match a predefined model described by a combined function of virtual inertia, damping and stiffness. In this paper, only pure stiffness rendering will be considered. To regulate the impedance perceived by a human, a model matching framework is employed as shown in Fig.~\ref{fig_impedance_matching}.

In the figure, $G_i$ is the open-loop model of the interactive system, also includes the open-loop impedance, $Z_d$ is the desired impedance, $\tau_d$ is the desired torque, $e$ is the torque tracking error, which also can be viewed as the impedance matching error, $d$ and $n$ are the disturbance and noise respectively, $G$ is the augmented open-loop model for the matching framework, $[\varphi_h,~d,~n]^T$ are the grouped external inputs, $[e,~u]^T$ are the outputs of interest to be optimized, and $y$ is the feedback signal flowing into the impedance controller $K$, which can be selected from the measured outputs or system states. Define the following transfer functions,
\begin{equation}
	G_3(s)=\frac{\tau_h(s)}{n(s)},~G_4(s)=\frac{\tau_h(s)}{d(s)}.
\end{equation}

Our objective is to synthesize an impedance controller $K$ so as to: (1) regulate the closed-loop impedance, which can smoothly vary in a wide range online; (2) strictly satisfy a number of frequency-domain constraints with respect to impedance rendering error, control effort and robustness; (3) obtain stable interaction via passivity.

\section{Constraints Characterization and Controller Design}
\label{Section3}

\subsection{Constraints Characterization}

The interaction between the human and robot usually falls into the low frequency range, and thus one only needs to restrict the impedance matching error, control effort and passivity in the low frequency bands. The disturbance may occur in the full frequency range, while the noise usually appears in a much high frequency range. These constraints can be determined according to the system's physical properties, and adjusted to achieve the desired performance by the designer.

Since the torque error $e$ directly indicates the impedance rendering accuracy, one can take the following constraint on the transfer function from $\varphi_h$ to $e$, i.e.,
\begin{equation}
\label{eq_req_e}
\left| {{T_{{\varphi _h}e}}(j\omega )} \right| \le {\gamma _1},~\left|\omega\right|  \le {\omega _e},
\end{equation}
which means the frequency response from $\varphi_h$ to $e$ should be bounded by the given scalar $\gamma_1>0$ in the desired low frequency range $\left[0~\omega_e\right]$.

To restrict the control effort within saturation limit, is has
\begin{equation}
\label{eq_req_u}
\left| {{T_{{\varphi _h}u}}(j\omega )} \right| \le {\gamma _2},~\left|\omega\right| \le {\omega _u}.
\end{equation}

Taking into consideration the disturbance and noise existing in the interactive system, one can consider
\begin{equation}
\label{eq_req_d}
\left| {{T_{d{\tau _h}}}(j\omega )} \right| \le {\gamma _3},~\omega  \in \mathcal{R},
\end{equation}
and
\begin{equation}
\label{eq_req_n}
\left| {{T_{n{\tau _h}}}(j\omega )} \right| \le {\gamma _4},~\left|\omega\right| \ge {\omega _n}.
\end{equation}

Passivity is a fundamental approach to guarantee rigorous stability in physical human-robot interaction applications. Some related definitions of passivity in both time-domain and frequency-domain are given as follows.
\begin{theorem}
	A stable single-input single-output system with input $u(t)$ and output $y(t)$ is said to be passive iff~\cite{Colgate1988}:
	\begin{equation}
		\int_{0}^{t}u(\tau)y(\tau)d\tau\ge E(t)-E(0)
	\end{equation}
	for all $t>0$.
\end{theorem}

In the above inequality, the integral part $\int_{0}^{t}u(\tau)y(\tau)d\tau$ is the total supplied energy flowing into the system. The difference between the final internal energy $E(t)$ and its initial energy $E(0)$ is the stored energy at time $t$. It can be inferred that at least some of the injected energy is dissipated from the system. Besides, when take $E(t)=0$, it has $-\int_{0}^{t}u(\tau)y(\tau)d\tau\le E(0)$, which reveals that a passive system can never deliver more energy than its initial to the outside. This inequality is a time-domain representation of passivity for both linear and nonlinear system, while the next one is in frequency-domain only for linear system.

\begin{theorem}
	A linear single-input single-output system with transfer function $G(s)$ is said to be passive iff~\cite{Anderson1972}:
	\begin{equation}
		\left| {\frac{{G(j\omega)}-1}{{G(j\omega)}+1}} \right| \le 1,~\omega\in \mathcal{R}.
	\end{equation}
\end{theorem}

It is clear that Theorem 2 is excessively conservative, since it is defined in the full frequency domain and fails to take into account that signals in physical systems usually fall into some restricted frequency ranges. Here, we will consider the following definition of the relaxed passivity.

\begin{theorem}
	A linear single-input single-output system with transfer function $G(s)$ is said to have relaxed passivity in the frequency range $|\omega|\le \omega_p$ iff~\cite{Haninger2018Tmech}:
	\begin{equation}
	\left| {\frac{{G(j\omega)}-1}{{G(j\omega)}+1}} \right| \le 1,~\left|\omega\right| \le {\omega _p}.
	\end{equation}
\end{theorem}

In this paper, the interactive velocity $\dot{\varphi}_h$ and interactive torque $\tau_h$ are taken as the input and output respectively for the passivity constraint. To guarantee stable interaction, one can consider the passivity constraint for the low frequency range to reduce its conservativeness, such that
\begin{equation}
\label{eq_passivity}
	\left| {\frac{{Z(j\omega)}-j\omega}{{Z(j\omega)}+j\omega}} \right| \le 1,~\left|\omega\right| \le {\omega _p}.
\end{equation}

Consequently, we have five frequency-domain constraints in (\ref{eq_req_e}), (\ref{eq_req_u}), (\ref{eq_req_d}), (\ref{eq_req_n}), (\ref{eq_passivity}) for the variable impedance control. The parameters $\gamma_{\{1, 2, 3, 4\}}>0$ and $\omega_{\{e, u, n, p\}}>0$ characterize the respective $H_\infty$ norm bounds at respective frequencies for each constraint, and can be tuned for interaction performance. Smaller bounds $\gamma_{\{1, 2, 3, 4\}}$ mean tighter restriction on the controller and better performance when a solution exists, while the frequency parameters $\omega_{\{e, u, n, p\}}$ are usually chosen to achieve the desired bandwidth and cutoff frequency.

\subsection{Scheduling of Controller Gains}

Without loss of generality, assume that the impedance controller $K(s)$ has $l \in \mathcal{N}^+$ inputs, such that
\begin{equation}
	K(s)=\left[K_1(s),~\ldots,~K_i(s),~\ldots,~K_l(s)\right],
\end{equation} 
with $i \in \left[1~l\right]$. The subcontroller $K_i(s)$ is assumed to be of the form
\begin{equation} 
	{K_i}(s) = \displaystyle \frac{K_{b_i0}s^m+\cdots+K_{b_ik}s^{m-k}+\cdots+K_{b_im}}
	{s^n+K_{a_i1}s^{n-1}+\cdots+K_{a_ij}s^{n-j}+\cdots+K_{a_in}},
\end{equation}
where, $n\in \mathcal{N}$ and $m\in \mathcal{N}$ are the orders of the denominator and numerator respectively, $K_i(s)$ is a proper transfer function with $m \leq n$, $j \in \left[1~n\right]$, $k \in \left[0~m\right]$.

The controller gains $K_{a_ij}$ and $K_{b_ik}$ are scheduled as a function of the desired impedance $Z_d$. Here, we interpolate the relationship with a polynomial function, such that
\begin{equation}
\label{eq_poly}
\left\{
\begin{aligned}
	K_{a_ij}(Z_d)&=K_{a_ij_0}+K_{a_ij_1}Z_d+\cdots+K_{a_ij_p}Z_d^p\\
	K_{b_ik}(Z_d)&=K_{b_ik_0}+K_{b_ik_1}Z_d+\cdots+K_{b_ik_p}Z_d^p\\
\end{aligned}
\right.,
\end{equation}
where, $p\in \mathcal{N}$ is the order of the polynomial. Coefficients $K_{a_ij_0},~\ldots,~K_{a_ij_p}$ and $K_{b_ik_0},~\ldots,~K_{b_ik_p}$ are tunable parameters to be determined.

There are several steps to determine those tunable parameters. Firstly, we need to select a finite number of design points $\{Z_{d1},~Z_{d2},~\ldots,~Z_{dN}\}$ which cover the range of the desired impedance $Z_d \in \left[0~Z_{d_{max}}\right]$. Secondly, we tune the controller gains $K_{a_ij}$ and $K_{b_ik}$ satisfying those performance constraints (\ref{eq_req_e})-(\ref{eq_passivity}) at each design point. Thirdly, we will fit the polynomial in (\ref{eq_poly}) using the resulting set of gains across design points, and obtain the coefficients $K_{a_ij_0},~\ldots,~K_{a_ij_p}$ and $K_{b_ik_0},~\ldots,~K_{b_ik_p}$. Finally, the controller gains corresponding to any desired impedance $Z_d \in \left[0~Z_{d_{max}}\right]$ can be obtained using the fitted polynomial (\ref{eq_poly}).

During the second step of tuning the controller gains, the $H_\infty$ synthesis technique can be employed. However, this multi-model, multi-frequency band and multi-objective optimization problem makes the conventional $H_\infty$ or LMI-based methods not applicable. Instead, a nonsmooth optimization method proposed in \cite{Apkarian2014} is adapted to find the solution and synthesize the controller. The nonsmooth optimization method has also demonstrated its validity and effectiveness in tuning gain-scheduled controllers for a three-loop autopilot application in \cite{Gahinet2013CDC}.

\section{Simulation and Experimental Results}
\label{Section4}

\begin{figure}[!htbp]
	\centering
	\includegraphics[width=0.5\columnwidth]{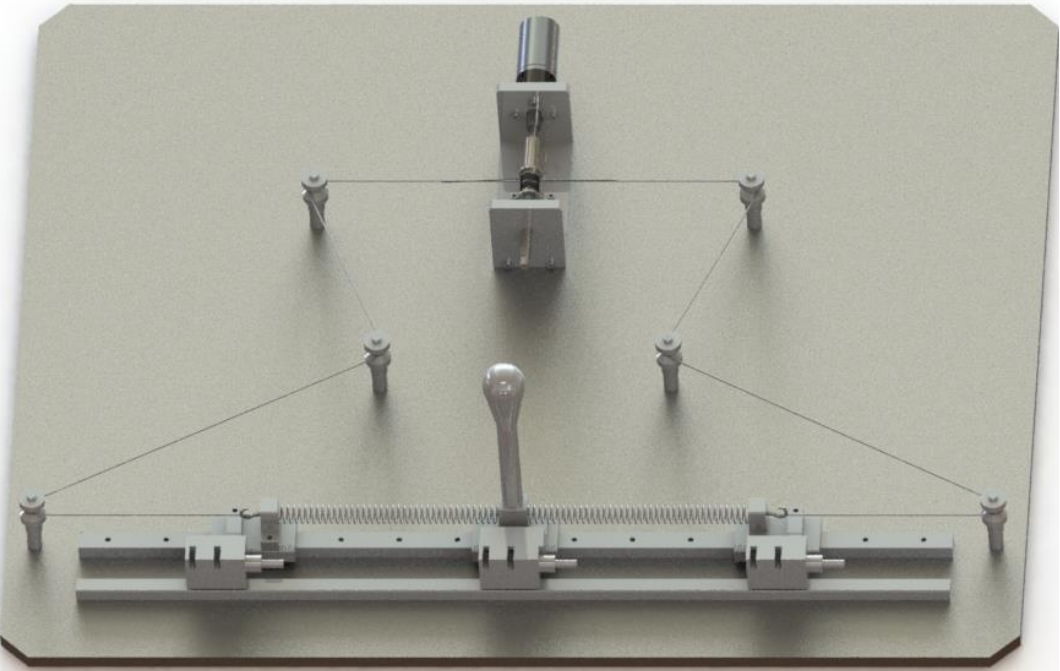}
	\caption{Mechanical design of the cable-driven SEA testbed.}
	\label{fig_Mechanical_Design}
\end{figure}

In this paper, a cable-driven series elastic actuator (SEA) system that has been used in our previous works \cite{Yu2017JAS,Yu2017AA,Yu2018MSSP,Yu2019CTA}, is used for simulation and experimental validation. Its mechanical design is shown in Fig. \ref{fig_Mechanical_Design}. The human can drive the handle moving along the guide, while the actuator regulates the actual impedance perceived by the human. 

\subsection{Simulations and Results}

The parameters involved in the performance constraints and gain scheduling are tunned to achieve good interaction performance, and listed in Table \ref{table_Nonsmooth_Variable_Parameters}. The actual interaction torque $\tau_h$ and torque error $e$ are chosen as the feedback signals, and grouped as the vector $y=\left[\tau_h,~e\right]^T$. To avoid control singularity, a weighting function $W_e(s)$ is appended after the torque error $e$ to create the weighted error $\tilde{e}$, i.e., $\tilde{e}=W_ee$. Detailed design procedure about the weighting function can be found in our previous work \cite{Yu2017AA}. The performance constraint (\ref{eq_req_e}) regarding the torque error will be replaced by 
\begin{equation}
\label{eq_req_weighted_e}
\left| {{T_{{\varphi _h} \tilde{e}}}(j\omega )} \right| \le {\gamma _1},~\left|\omega\right|  \le {\omega _e}.
\end{equation}

\begin{table}[!htbp]
	\caption{Parameters for the variable impedance control}
	\label{table_Nonsmooth_Variable_Parameters}
	\begin{center}
		\renewcommand{\arraystretch}{1.25}		
			\begin{tabular}{cc|cc}
				\hline \hline	
				Parameter & Value & Parameter & Value \\
				\hline
				$\gamma_1$ & 0.05 & $\gamma_2$ & 44 \\
				\hline
				$\gamma_3$ & 0.03 & $\gamma_4$ & 0.3 \\
				\hline
				$\omega_e$ & $12\pi$ & $\omega_u$ & $12\pi$ \\
				\hline
				$\omega_n$ & $40\pi$ & $\omega_p$ & $12\pi$ \\
				\hline
				$l$ & $2$ & $n$ & 1\\
				\hline
				$m$ & 1 & $p$ & 5\\
				\hline \hline
		\end{tabular}
	\end{center}
\end{table}

\begin{figure}[!htbp]
	\centering
	\includegraphics[width=0.7\columnwidth]{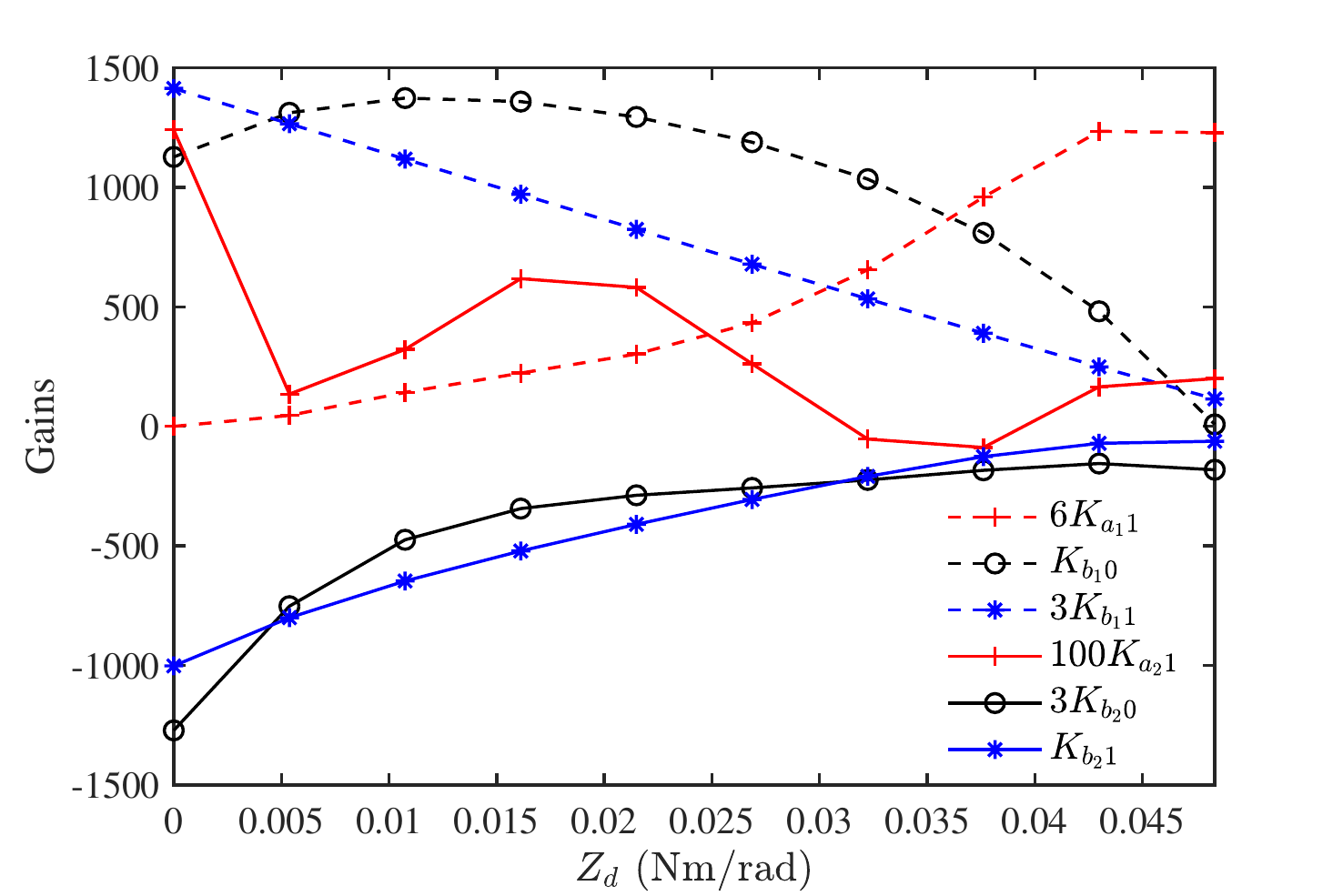}
	\caption{Controller gains versus the desired stiffness at each design point. The gains are multiplied by respective factors to fit into one figure.}
	\label{fig_Gain_Stiffness}
\end{figure}

We evenly chose 10 design points in the stiffness range $Z_d \in \left[0~K_s\right]$, where, $K_s$ is the physical stiffness of the SEA. Then, the nonsmooth optimization is employed to parameterize the polynomial (\ref{eq_poly}) for each controller gain $K_{a_11}$, $K_{b_10}$, $K_{b_11}$ and $K_{a_21}$, $K_{b_20}$, $K_{b_21}$. The controller gains at the 10 design points are presented in Fig. \ref{fig_Gain_Stiffness}, which shows that the gains can be smoothly fitted with the polynomials.

\begin{figure}[!htbp]
	\centering
	\includegraphics[width=0.6\columnwidth]{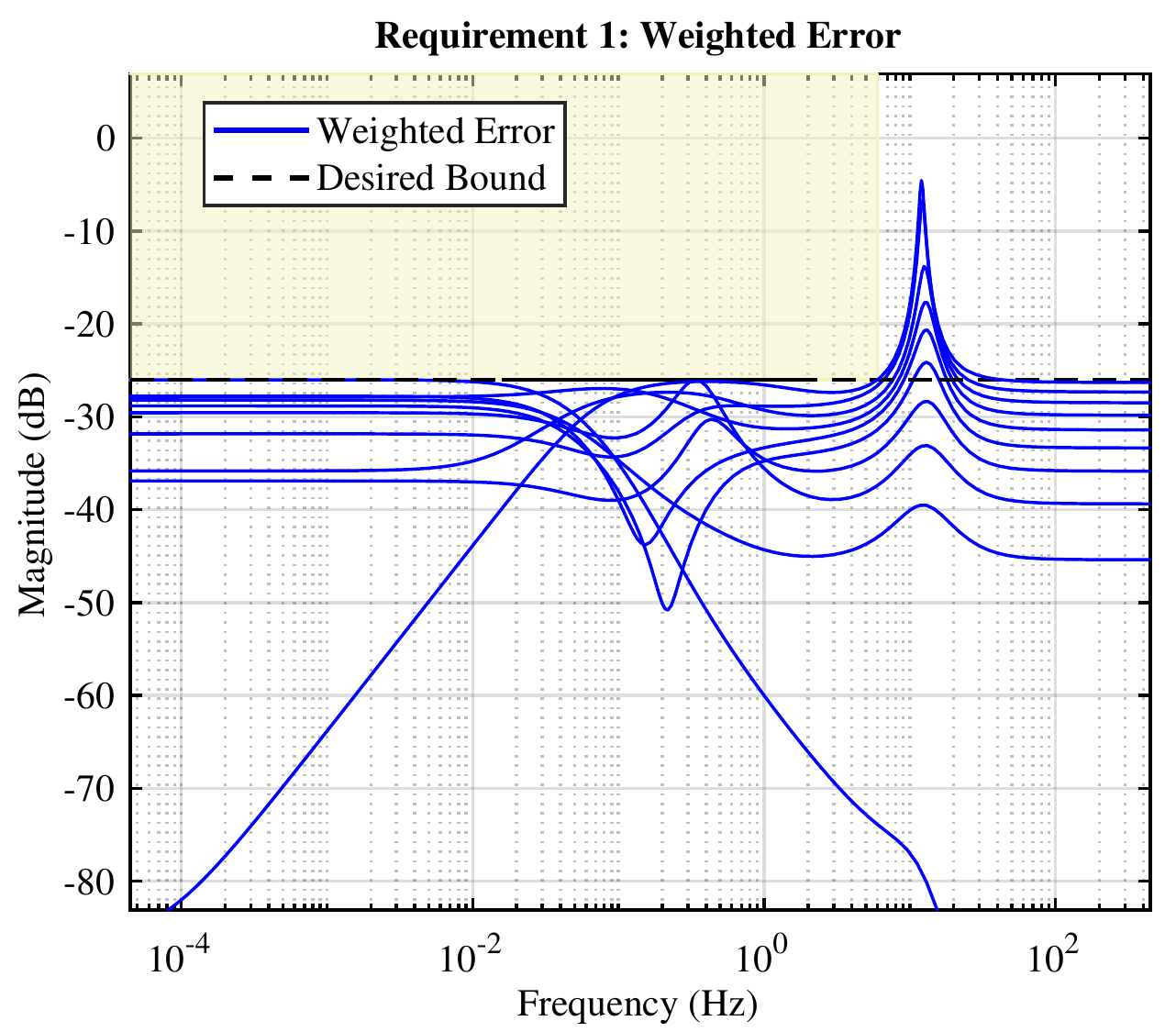}
	\caption{Simulated frequency responses $\left| {{T_{{\varphi _h}\tilde e}}(j\omega )} \right|$ for each design point.}
	\label{fig_Simulation_Responses_Error}
\end{figure}

\begin{figure}[!htbp]
	\centering
	\includegraphics[width=0.6\columnwidth]{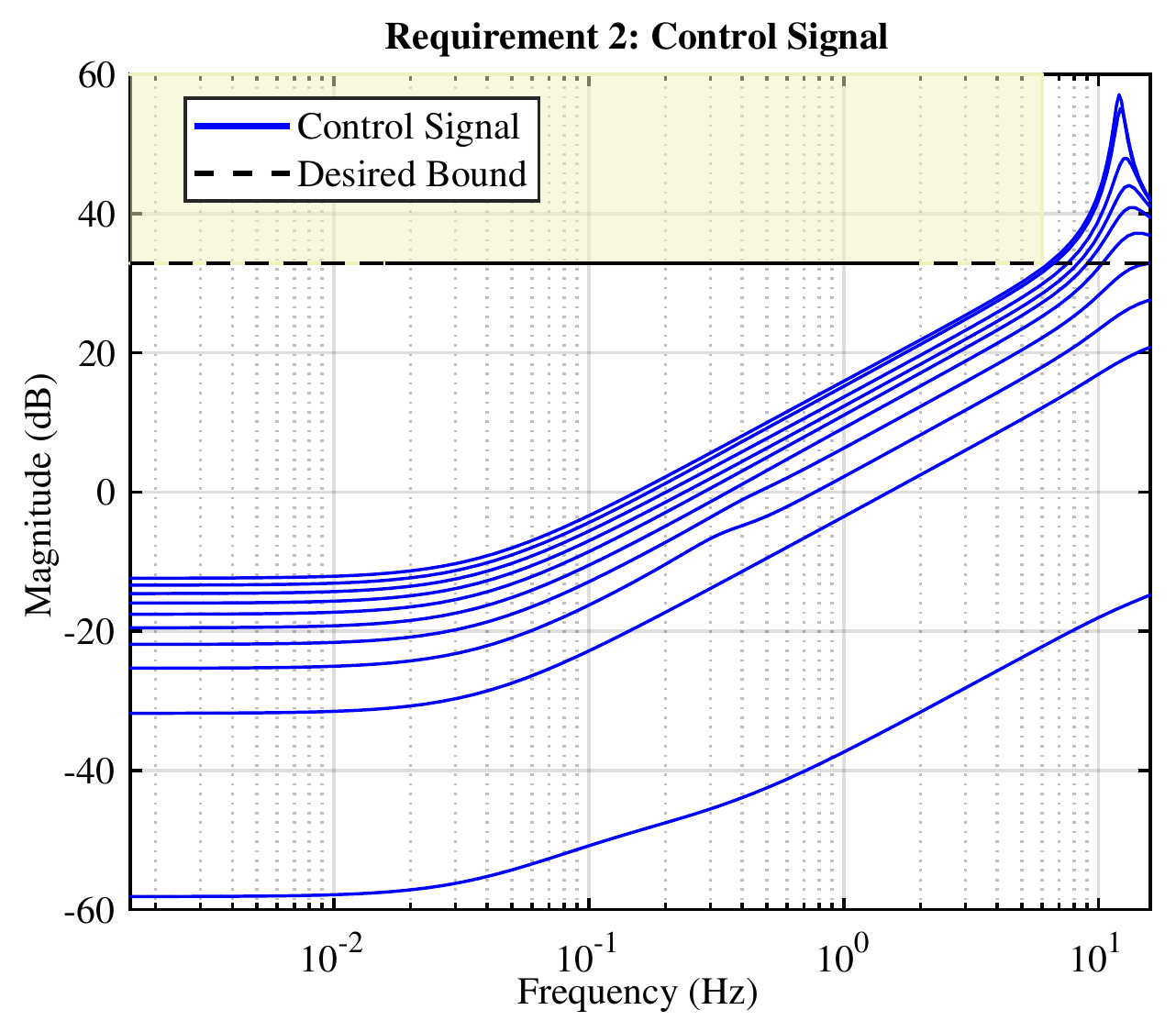}
	\caption{Simulated frequency responses $\left| {{T_{{\varphi _h}u}}(j\omega )} \right|$ for each design point.}
	\label{fig_Simulation_Responses_Control_Signal}
\end{figure}

\begin{figure}[!htbp]
	\centering
	\includegraphics[width=0.6\columnwidth]{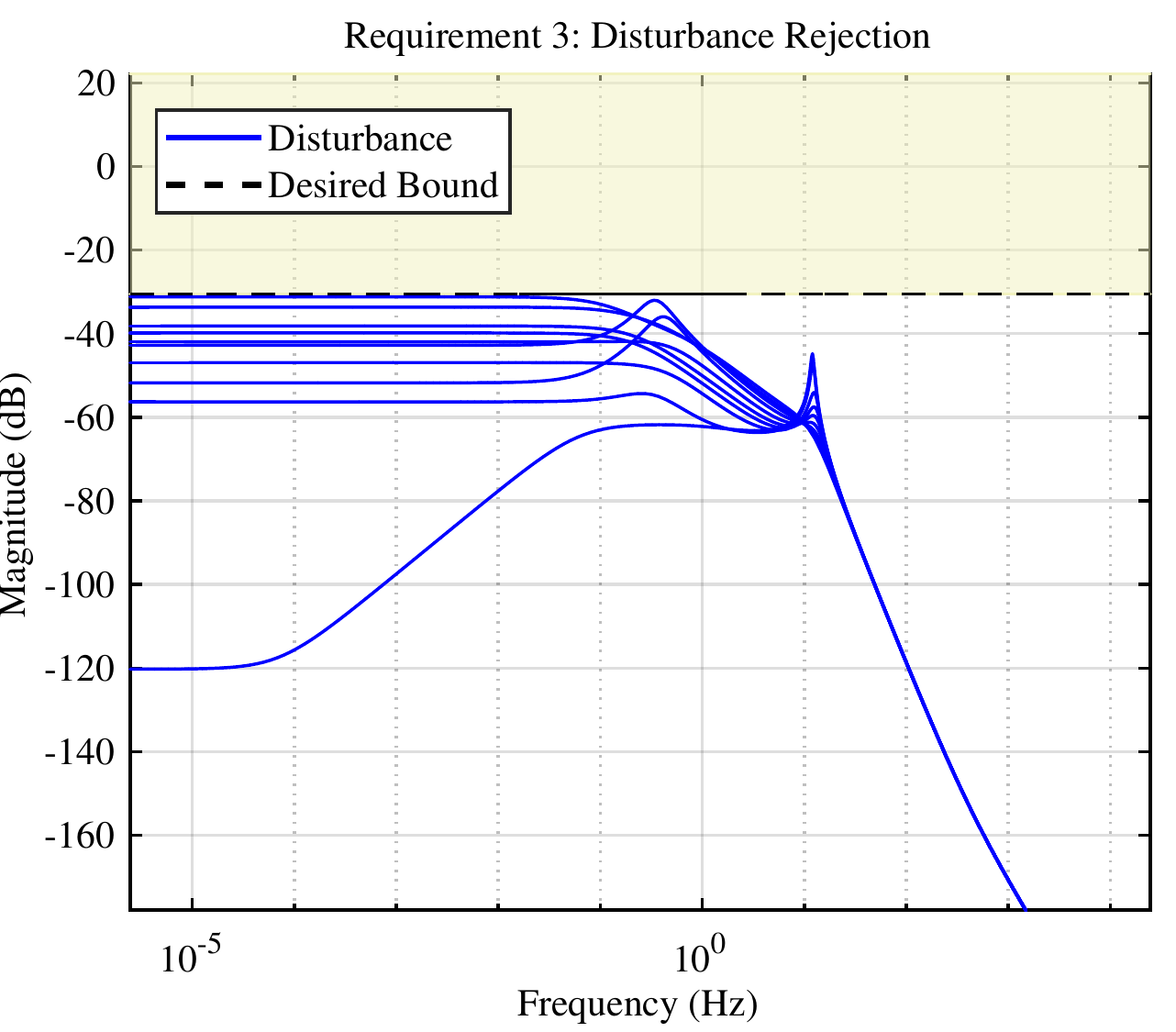}
	\caption{Simulated frequency responses $\left| {{T_{d{\tau _h}}}(j\omega )} \right|$ for each design point.}
	\label{fig_Simulation_Responses_Disturbance}
\end{figure}

\begin{figure}[!htbp]
	\centering
	\includegraphics[width=0.6\columnwidth]{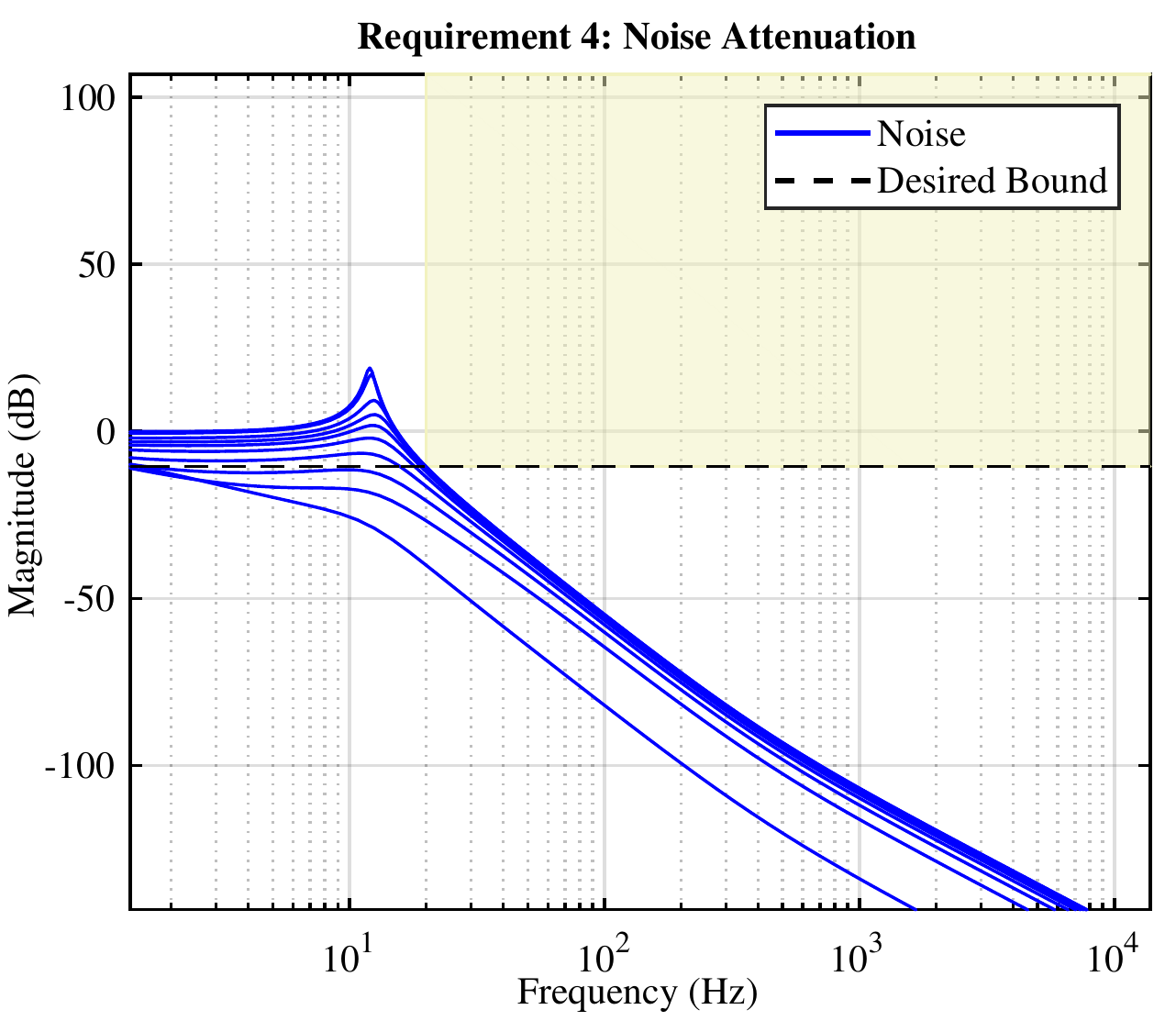}
	\caption{Simulated frequency responses $\left| {{T_{n{\tau _h}}}(j\omega )} \right|$ for each design point.}
	\label{fig_Simulation_Responses_Noise}
\end{figure}

\begin{figure}[!htbp]
	\centering
	\includegraphics[width=0.6\columnwidth]{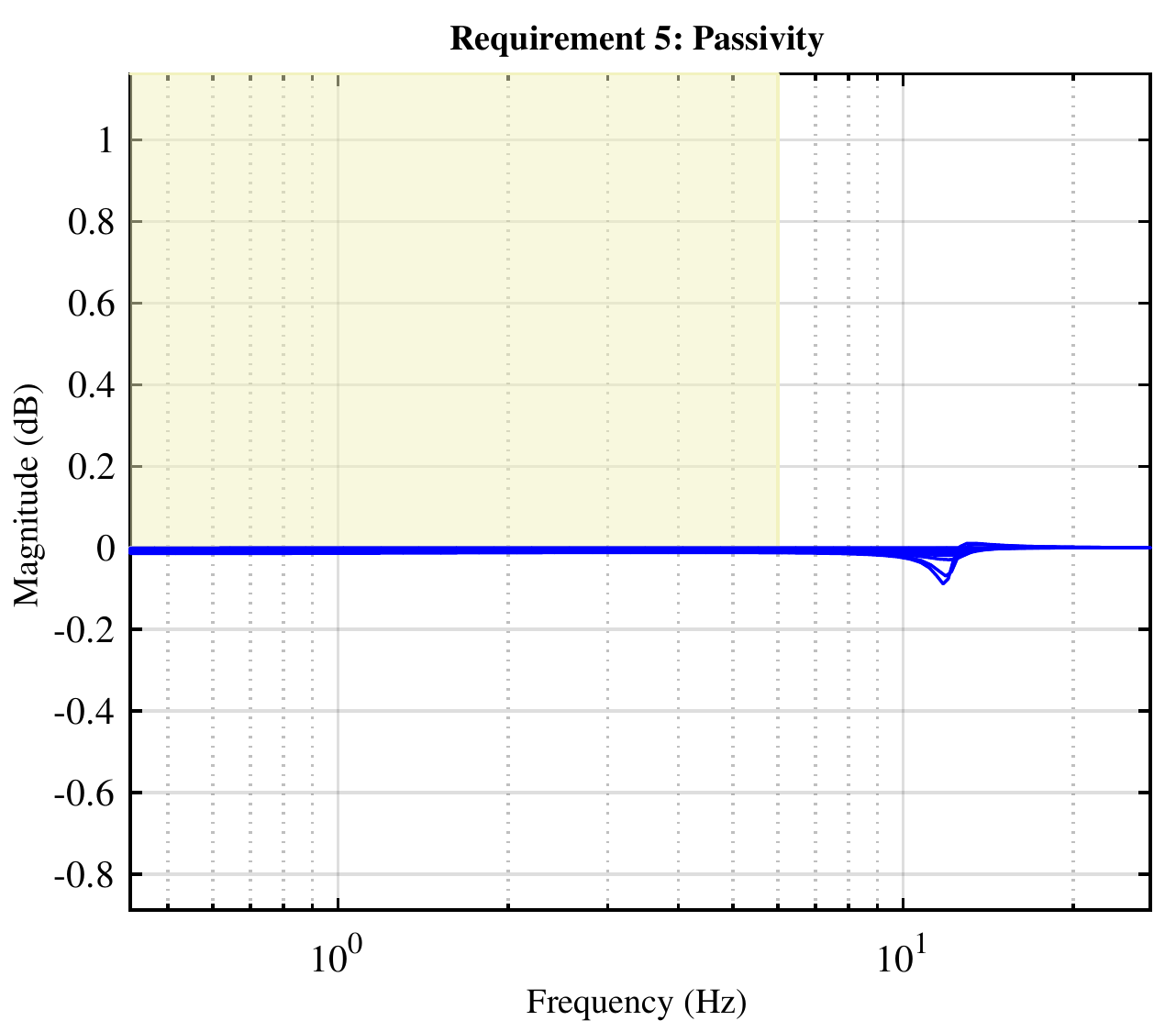}
	\caption{Simulated frequency responses $\left| {\frac{{Z(j\omega)}-j\omega}{{Z(j\omega)}+j\omega}} \right|$ for each design point.}
	\label{fig_Simulation_Responses_Passivity}
\end{figure}

To check whether the performance constraints are all satisfied, the frequency responses about the weighted error, control signal, disturbance attenuation, noise rejection and passivity, are illustrated in Fig. \ref{fig_Simulation_Responses_Error}-\ref{fig_Simulation_Responses_Passivity}. At all those design points, the performance constraints can be guaranteed within the given bounds at the specified frequency ranges.

\begin{figure}[!htbp]
	\centering
	\subfigure{\includegraphics[width=0.7\columnwidth]{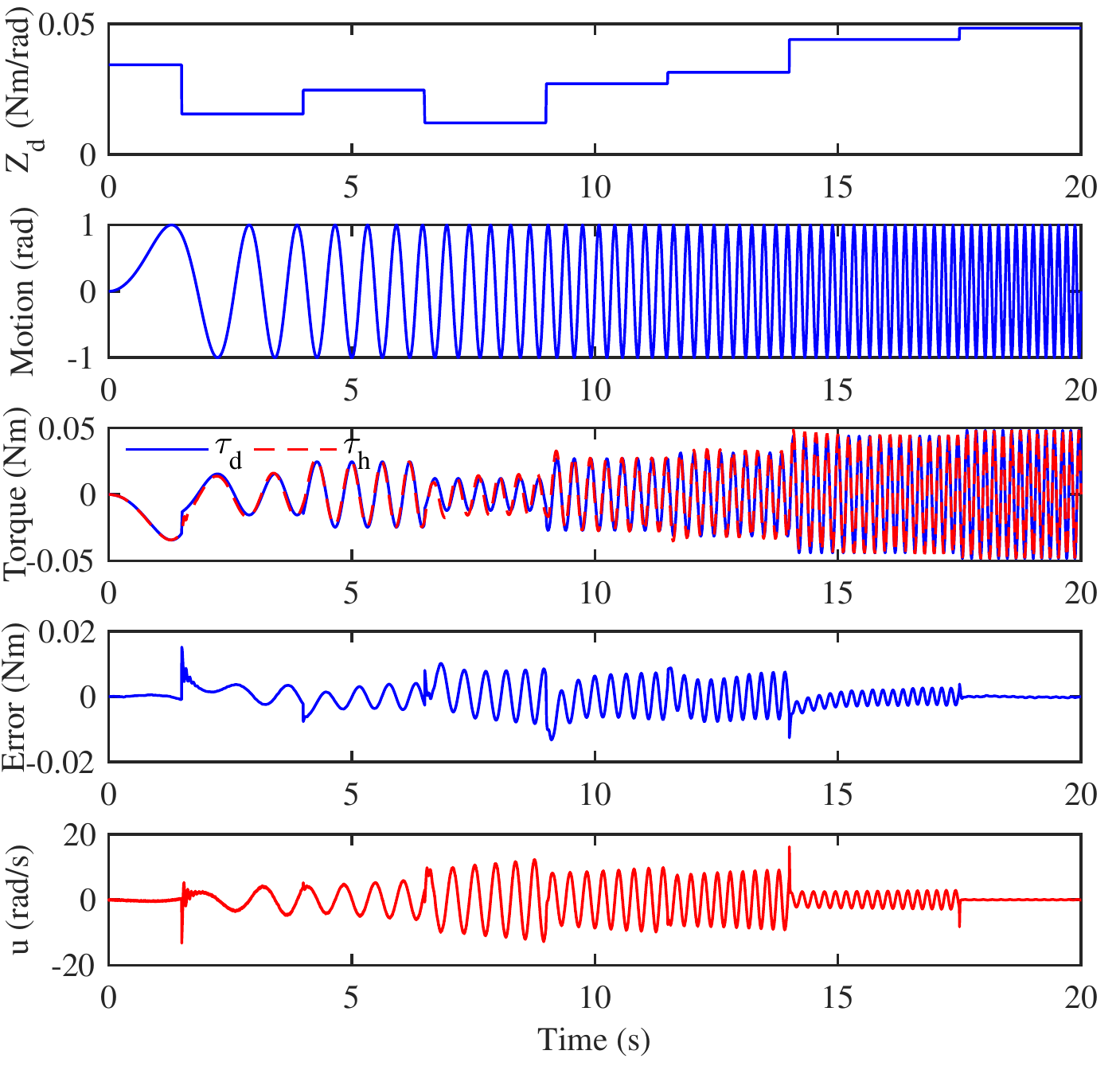}}
	\subfigure{\includegraphics[width=0.7\columnwidth]{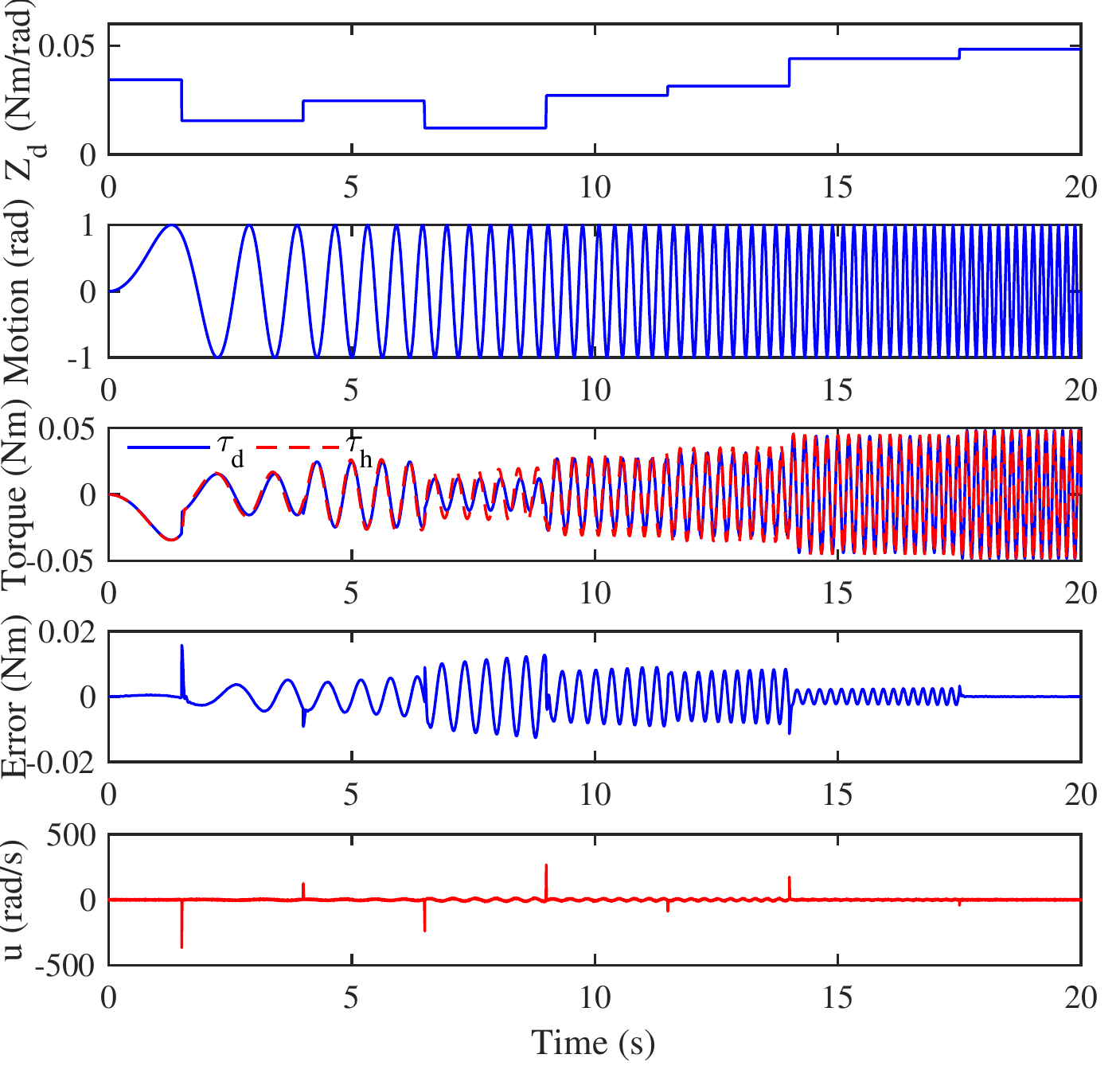}}
	\caption{Simulation results of variable stiffness control. Top: gain-scheduled method. Bottom: gain-fixed PID method.}
	\label{fig_sim}
\end{figure}

To further demonstrate the performance, simulations of variable impedance control are conducted and compared with a gain-fixed passivity-based PID method. The parameters of the PID are the same as in our previous work \cite{Yu2018MSSP}, and  remain constant during the variable impedance control. During the simulations, the desired stiffness changes according to the sequence $\{0.71K_s,$ $0.32K_s,$ $0.51K_s,$ $0.25K_s,$ $0.56K_s,$ $0.65K_s,$ $0.91K_s,$ $K_s\}$. The human hand motion is set as a chirp signal with frequency changing from 0 to 6 Hz. The simulation results of the two methods, including the desired stiffness $Z_d$, human hand motion $\varphi_h$, desired interaction torque $\tau_d$, actual interaction torque $\tau_h$, torque error $e$, control signal $u$, are shown in Fig. \ref{fig_sim}.

To better compare the two methods, the maximal torque error (ME), sum of squared error (SSE), maximal controller output (MCO), and signal to noise ratio of the controller output (SNR) are calculated and summarized in Table \ref{Table_Performance_Comparison_Sim}. The proposed gain-scheduled method obtains more accurate impedance control because of its smaller ME and SSE compared with the PID method. Its maximal controller output is only 16.3 rad/s below the motor saturation limit (44 rad/s), while the PID method is 367.4 rad/s far above the limit. Besides, its signal to noise ratio is also much higher than the PID method. Thus, during the variable stiffness control, the proposed gain-scheduled method achieves more accurate stiffness shaping, more smoother stiffness transition, and more robust disturbance attenuation and noise rejection.

\begin{table}[!htbp]
	\caption{Simulation results: comparison between the two methods}
	\label{Table_Performance_Comparison_Sim}
	\begin{center}
		\renewcommand{\arraystretch}{1.25}
		\begin{tabular}{l|cc}
			\hline \hline	
			& Gain-scheduled & PID \\
			\hline
			ME (Nm) & 0.0152 & 0.0158 \\
			\hline
			SSE ((Nm)$^2$) & 0.5484 & 0.8002 \\
			\hline
			MCO (rad/s) & 16.3 & 367.4 \\
			\hline
			SNR (dB) & 21.1 & 0.9 \\
			\hline \hline
		\end{tabular}
	\end{center}
\end{table}

\subsection{Experiments and Results}

\begin{figure}[!htbp]
	\centering
	\subfigure{\includegraphics[width=0.72\columnwidth]{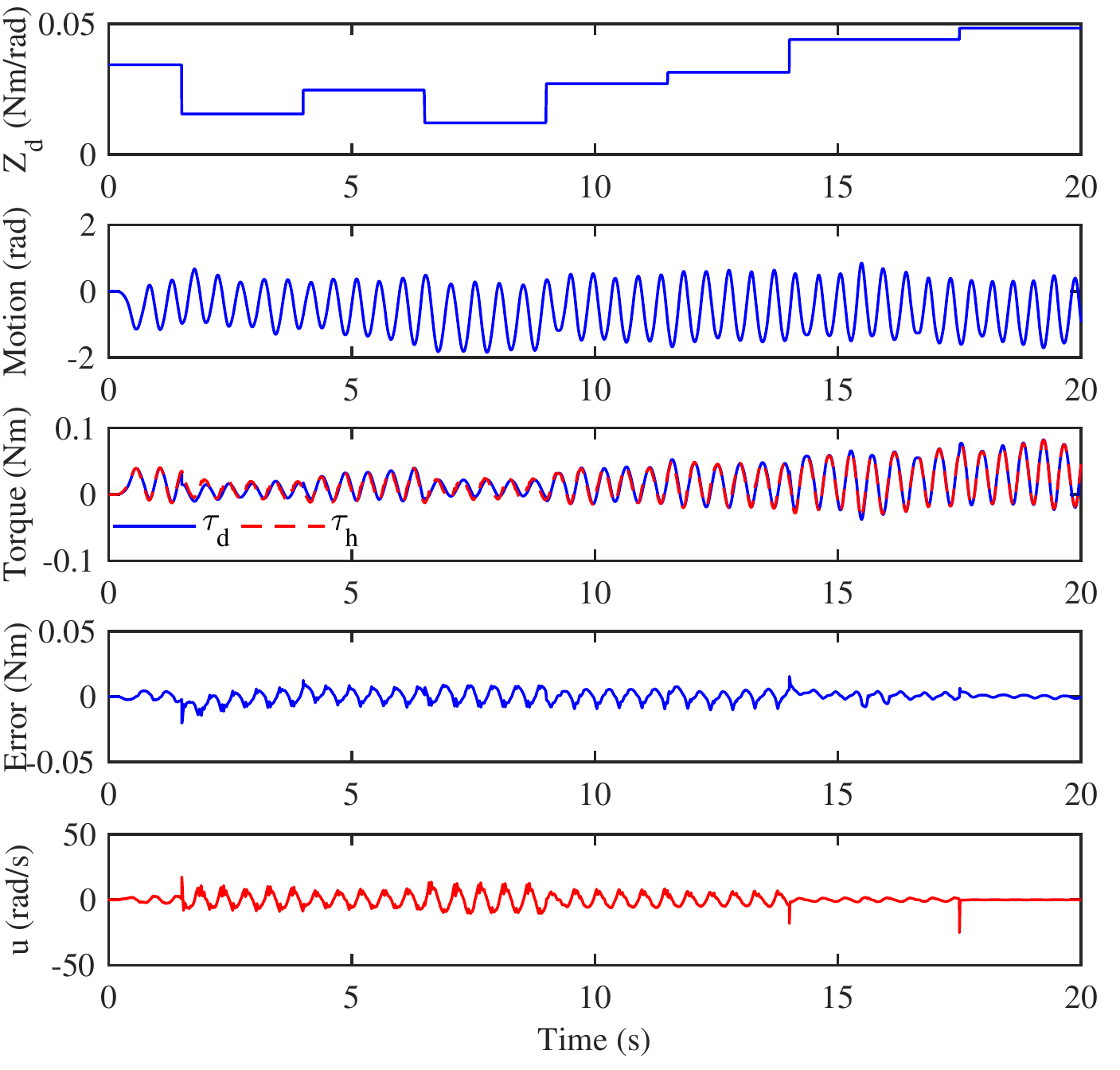}}
	\subfigure{\includegraphics[width=0.72\columnwidth]{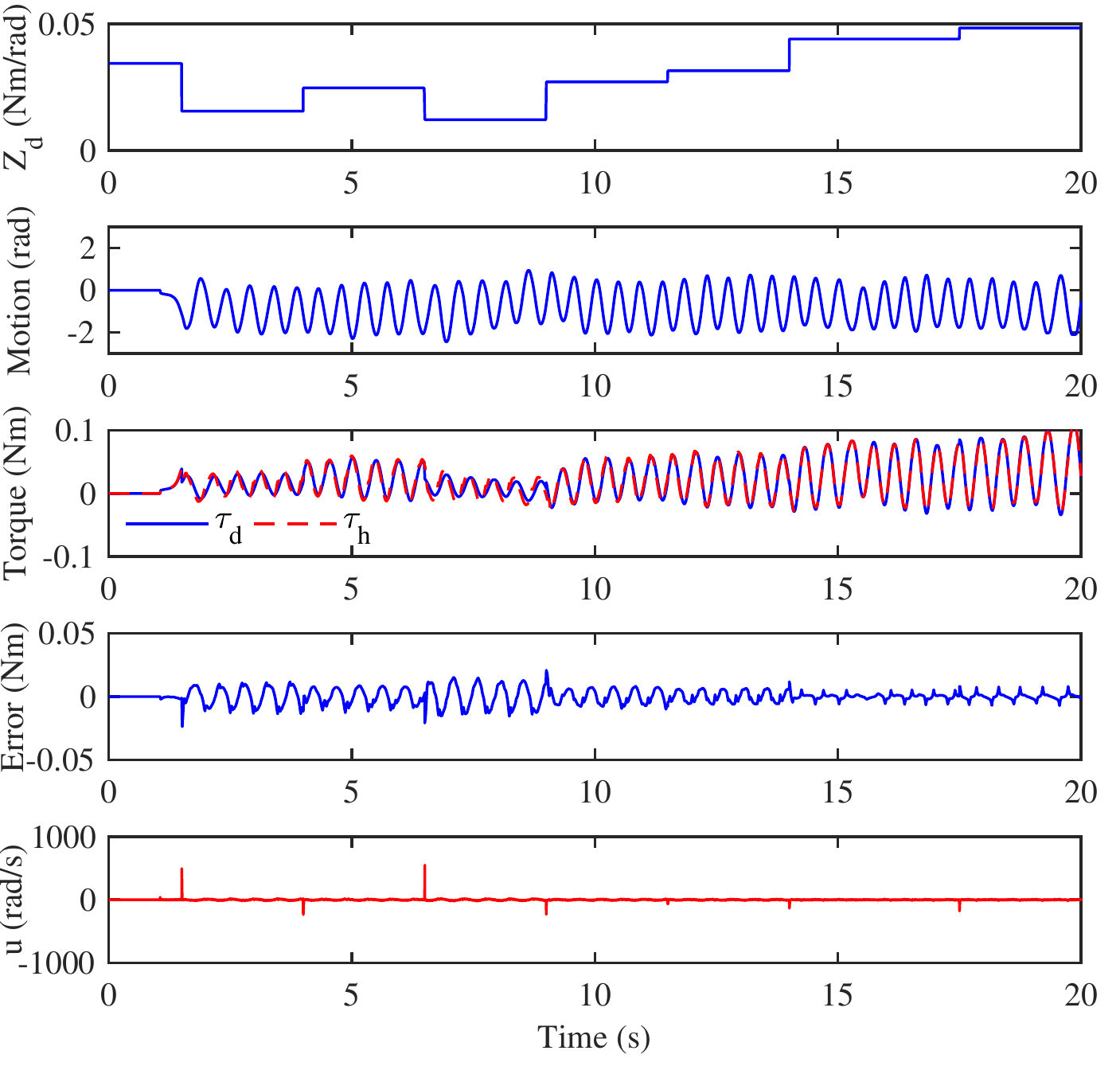}}
	\caption{Experimental results of variable stiffness control. Top: gain-scheduled method. Bottom: gain-fixed PID method.}
	\label{fig_exp}
\end{figure}

Experiments are also conducted to verify its performance. The desired stiffness changes according to the same sequence in the simulations. The motion $\varphi_h$ will be applied by the human hand. The experimental results of the two methods are presented in Fig. \ref{fig_exp}, and the quantified metrics are also calculated and listed in Table \ref{Table_Performance_Comparison_Exp}.

\begin{table}[!htbp]
	\caption{Experimental results: comparison between the two methods}
	\label{Table_Performance_Comparison_Exp}
	\begin{center}
		\renewcommand{\arraystretch}{1.25}
		\begin{tabular}{l|cc}
			\hline \hline	
			& Gain-scheduled & PID \\
			\hline
			ME (Nm) & 0.0203 & 0.0236 \\
			\hline
			SSE ((Nm)$^2$) & 0.6613 & 1.2119 \\
			\hline
			MCO (rad/s) & 25.2 & 548.7 \\
			\hline
			SNR (dB) & 12.0 & 2.8 \\
			\hline \hline
		\end{tabular}
	\end{center}
\end{table}

From the experimental results, the proposed gain-scheduled method still achieves more accurate and robust stiffness matching, and well switches from one desired stiffness to another. By guaranteeing the passivity in the low frequency range, stability of the coupled human-robot system is ensured.

\section{Conclusion}
\label{Section5}

This paper proposes a gain-scheduled variable stiffness control method for physical human-robot interaction, with the performance constraints restricted into respective frequency ranges. Especially, the passivity constraint is relaxed in such a way that it is only required in the low frequency range, and thus the controller is less conservative. This relaxation is reasonable due to the restricted frequency properties of the interactive robots. The controller gains are scheduled with respect to the desired stiffness, and the scheduling function is tunned by a nonsmooth optimization method. Compared with the gain-fixed passivity-based PID method, the proposed approach achieves more accurate and robust impedance rendering, and smoother stiffness transition in both simulations and experiments.

Future work will be focused on the controller design and rigorous analysis of passivity for general time-varying variable impedance/admittance control and high-DOFs interactive robots.

\bibliography{reference}

\end{document}